\documentclass{article}
\usepackage{graphicx}
\usepackage{amsmath}

\usepackage{subcaption}

\usepackage[final,main, nonatbib]{neurips_2025}

\usepackage[utf8]{inputenc} 
\usepackage[T1]{fontenc}    
\usepackage{hyperref}       
\usepackage{url}            
\usepackage{booktabs}       
\usepackage{amsfonts}       
\usepackage{nicefrac}       
\usepackage{microtype}      
\usepackage{xcolor}         
\usepackage[backend=biber,style=numeric]{biblatex}
\title{Improving Sparse Memory Finetuning}

\author{%
  Satyam Goyal, Anirudh Kanchi, Garv Shah, Prakhar Gupta \\
  University of Michigan, Ann Arbor\\
  \texttt{sagoyal@umich.edu, akanchi@umich.edu, garvshah@umich.edu, prakharg@umich.edu} \\
}

\begin{document}

\maketitle
\begin{abstract}

Large Language Models (LLMs) are typically static after training, yet real-world applications require continual adaptation to new knowledge without degrading existing capabilities. Standard approaches to updating models, like full finetuning or parameter-efficient methods (e.g., LoRA), face a fundamental trade-off: catastrophic forgetting. They modify shared dense representations, causing interference across tasks. Sparse Memory Finetuning (SMF) offers a promising alternative by localizing updates to a small subset of parameters in explicit memory layers. In this work, we present an open-source pipeline to retrofit existing pretrained models (Qwen-2.5-0.5B) with sparse memory modules, enabling effective continual learning on consumer hardware. We extend prior work by introducing a theoretically grounded slot-selection mechanism based on Kullback-Leibler (KL) divergence, which prioritizes memory updates for informationally "surprising" tokens relative to a background distribution. Our experiments demonstrate that our retrofitted models can acquire new factual knowledge with minimal forgetting of held-out capabilities, validating the sparse update hypothesis in a practical setting.

\end{abstract}

\section{Introduction}

Modern Large Language Models (LLMs) are typically trained once on a massive corpus and then deployed as static artifacts. However, the information they model is dynamic: new events occur, policies change, and domain-specific knowledge evolves. The practical goal of \textit{continual learning} is to incorporate this new information efficiently while preserving the model's previously learned capabilities. The primary obstacle is \textit{catastrophic forgetting} \cite{luo2023empirical}, where the gradient updates required to learn distribution-shifted data degrade performance on older tasks.

Standard mitigation strategies often fall short in the "lifelong" regime. Replay buffers \cite{chaudhry2019tiny} are data-inefficient and raise privacy concerns. Parameter-Efficient Finetuning (PEFT) methods, such as Low-Rank Adaptation (LoRA) \cite{hu2022lora}, significantly reduce the number of trainable parameters but do not fundamentally solve the interference problem; because the low-rank updates are applied densely to the hidden states, a single parameter update can still impact global model behavior. A complementary line of work suggests augmenting models with explicit memory---either non-parametric retrieval (RAG) \cite{lewis2020retrieval} or learned memory components \cite{sukhbaatar2015end}---to reduce the necessity of overwriting synaptic weights. However, these solutions introduce high memory and cost overheads that can't scale unbounded.

In this work, we build upon the recent proposal of Sparse Memory Finetuning (SMF) \cite{lin2025continuallearningsparsememory}, which hypothesizes that interference can be minimized by restricting updates to a sparse subset of parameters explicitly tied to the incoming data. While promising, prior implementations rely on rigid heuristics (e.g., TF-IDF) for memory slot selection and are often tied to proprietary or custom model architectures. We address these limitations by developing a generalizable pipeline for retrofitting standard open-weights Transformers with sparse memory layers.

Our primary contributions are threefold:
\begin{enumerate}
    \item \textbf{Open-Source Retrofitting Pipeline:} We provide a reproducible methodology to surgically replace Feed-Forward Networks (FFNs) in pretrained Transformers (specifically Qwen-2.5-0.5B) with sparse key-value memory layers, followed by a "healing" stage to recover general capability.
    \item \textbf{Information-Theoretic Slot Selection:} We critique the standard TF-IDF heuristic for identifying trainable memory slots and propose a novel scoring rule based on KL-divergence. This method selects slots based on the information gain of the current batch relative to a background usage distribution, providing a more principled signal for sparsity.
    \item \textbf{Empirical Validation of Plasticity-Stability Trade-off:} We demonstrate that our retrofitted models can learn new tasks (TriviaQA) via sparse updates while maintaining higher stability on held-out benchmarks (GSM8k, NaturalQuestions) compared to dense finetuning baselines.
\end{enumerate}


\section{Related Work}

The primary challenge in continual learning is preventing the degradation of prior knowledge while acquiring new information. Classical approaches include regularization techniques to constrain weight updates \cite{luo2023empirical} and replay-based methods that interleave old data with new \cite{chaudhry2019tiny}. While replay is effective, it requires maintaining a buffer of sensitive training data and incurs significant compute overhead.

PEFT methods like LoRA \cite{hu2022lora} and Adapters freeze the backbone and inject a small number of trainable parameters. While efficient, LoRA operates by adding low-rank matrices to dense layers. As noted by \cite{lin2025continuallearningsparsememory}, LoRA updates are global: modifying the low-rank adapters affects the representation of all tokens in the embedding space. This results in significant interference when tasks vary in distribution. 

Retrieval-Augmented Generation (RAG \cite{RAG}) mitigates forgetting by storing knowledge in external non-parametric memory, enabling rapid updates without modifying model weights, though it relies heavily on retrieval quality. 
While powerful, RAG relies on the quality of a fixed retriever and introduces latency and complexity at inference time.

\textit{Parametric} approaches, such as Memory Networks \cite{sukhbaatar2015end}, integrate memory directly into the network weights. Memory-R1 (\cite{yan2025memoryr1enhancinglargelanguage}) introduces persistent memory layers that accumulate task-specific representations over time, while SPARC (\cite{pca}) leverages sparse activation and selective parameter updates to enable continual adaptation with minimal interference. Recent work has scaled this concept to deep Transformers \cite{berges2025memory}, demonstrating that sparse memory layers can replace FFNs without loss of pretraining performance. These approaches insert dedicated memory layers or key–value memory banks at multiple depths, enabling models to store and retrieve task-specific representations while keeping the backbone largely frozen. By updating only a small subset of memory parameters per task, they scale to long task sequences with minimal interference, constant-time adaptation costs, and reduced catastrophic forgetting.

Our work extends this by investigating the \textit{dynamic} properties of these layers during finetuning, specifically comparing heuristic slot selection (TF-IDF) against information-theoretic approaches (KL-Divergence). 

\section{Method}

\subsection{The Standard Feed-Forward Network}
In a standard Transformer architecture, the Feed-Forward Network (FFN) at layer $l$ processes the output of the attention mechanism. Given an input token representation $x \in \mathbb{R}^{d}$, the FFN typically consists of two dense linear transformations separated by a non-linearity $\sigma$:
\begin{equation}
    \text{FFN}(x) = W_2 \cdot \sigma(W_1 x)
\end{equation}
where $W_1 \in \mathbb{R}^{d_{\text{ff}} \times d}$ and $W_2 \in \mathbb{R}^{d \times d_{\text{ff}}}$. While effective, this operation is dense: every parameter in $W_1$ and $W_2$ contributes to the processing of every token. Consequently, an update to any weight $\Delta W$ potentially alters the model's behavior for all future inputs, creating a high risk of catastrophic interference during sequential learning.

\subsection{Sparse Memory Layers}
To enable interference-free updates, we adopt the \textbf{Memory Layer} architecture \cite{lin2025continuallearningsparsememory, berges2025memory}. A memory layer replaces the dense FFN with a sparse key-value lookup mechanism. It consists of a query projector $W_q \in \mathbb{R}^{d \times d}$, a set of trainable keys $K \in \mathbb{R}^{M \times d_k}$, and a set of trainable values $V \in \mathbb{R}^{M \times d}$, where $M$ is the memory size (number of slots).

For an input $x$, the layer generates a query $q = W_q x$. It then retrieves the indices $\mathcal{I}$ of the top-$k$ keys maximizing the inner product with $q$:
\begin{equation}
    \mathcal{I} = \text{Top-k}(\{q \cdot k_i\}_{i=1}^M)
\end{equation}
The output is a weighted sum of the retrieved values, typically gated and added to the residual stream:
\begin{equation}
    y = x + \alpha \sum_{i \in \mathcal{I}} p_i v_i, \quad p_i = \text{Softmax}(q \cdot k_i)_{i \in \mathcal{I}}
\end{equation}
Crucially, forward propagation only activates $k$ out of $M$ slots (where $k \ll M$). This sparsity structure implies that gradient updates can be localized: if we only update the values $v_i$ for $i \in \mathcal{I}$, parameters associated with un-accessed slots remain frozen, theoretically preserving knowledge stored in those regions.

\begin{figure}[t]
    \centering
    \includegraphics[width=0.4\linewidth]{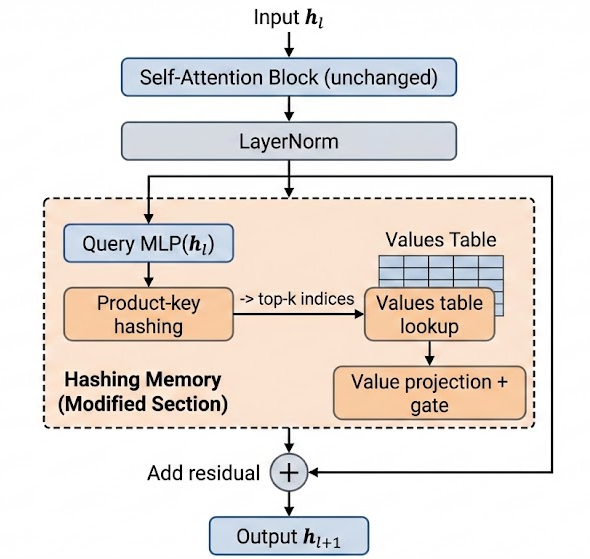}
    \caption{Retrofitted pretrained Transformer with sparse Memory Layers.}
    \label{fig:retrofit}
\end{figure}

\subsection{Three-Stage Retrofit Pipeline}
Our approach retrofits a dense pretrained Large Language Model (LLM) into a sparse memory-augmented model capable of continual learning. We describe the three-stage pipeline and the specific slot-selection algorithms used to minimize interference:

\paragraph{Stage 1: Retrofitting.}
We start from \textbf{Qwen-2.5-0.5B-Instruct} and replace a small set of FFN layers with initialized memory layers (layers $[8,12,16]$ based on analysis by \cite{lin2025continuallearningsparsememory}). Immediately after replacement, model behavior degrades because the forward computation has changed. The original dense weights ($W_{up}, W_{down}, W_{gate}$) for these layers are discarded, and new sparse memory modules (Keys $K$, Values $V$) are initialized. This drastic change initially degrades model perplexity, necessitating a recovery phase.

\paragraph{Stage 2: Recovery (Healing).}
To restore baseline competence, we finetune only the new memory parameters on a general instruction dataset. We use 20,000 samples from OpenAssistant (oasst1). This phase aligns the random memory projections with the pretrained residual stream, ensuring the model can produce coherent text before learning new tasks; the purpose is not to learn a new task but to adapt the memory layers.

\begin{figure}[t]
    \centering
    \includegraphics[width=0.6\linewidth]{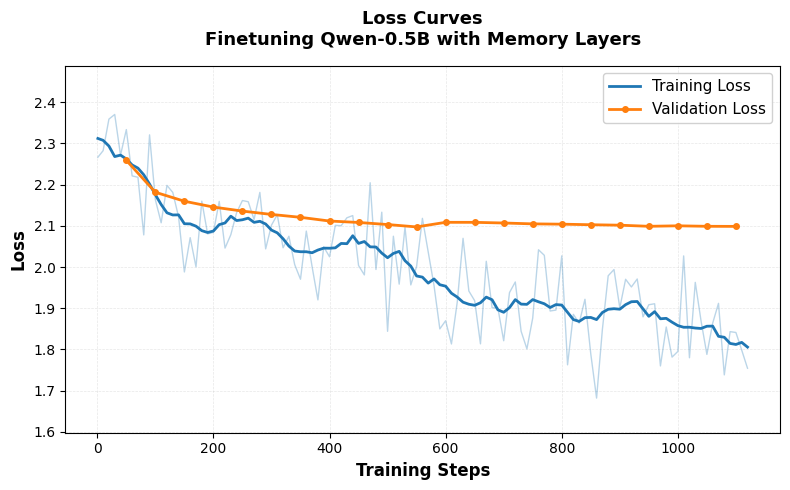}
    \caption{Recovery phase training loss on the instruction dataset (using TF-IDF).}
    \label{fig:recovery}
\end{figure}

\paragraph{Stage 3: Task-Specific Finetuning.}
We then finetune on the target task (HellaSwag, 1k samples) using two alternatives: \textbf{(i) full/dense finetuning} (updates dense weights) and \textbf{(ii) sparse memory finetuning} (updates only a small subset of memory entries per batch, keeping the base model frozen).

\subsection{Sparse Update via Gradient Masking}
The key mechanism in SMF is that even within a memory table, we avoid updating all entries. During each forward pass, the memory layer touches only indices $\mathcal{I}$ (the retrieved slots). We implement sparse updating by masking gradients so that only rows corresponding to selected indices receive non-zero gradient. In the simplest form, this becomes:
\begin{equation}
\theta_{\text{new}} = \theta_{\text{old}} - \eta \, \nabla_\theta \mathcal{L} \cdot \mathbb{I}[i \in \mathcal{I}],
\end{equation}
where $\theta$ are the memory parameters (e.g., value table rows) and $\mathbb{I}[i \in \mathcal{I}]$ is a binary mask.


Operationally, our implementation logs which memory indices are accessed during the forward pass, constructs a boolean mask over memory rows, and registers a gradient hook on the memory value matrix so that gradients for unselected rows are zeroed before the optimizer step. This ensures unused memory slots remain unchanged, which is the intended mechanism for reducing interference.

\begin{figure}[t]
    \centering
    \includegraphics[width=0.7\linewidth]{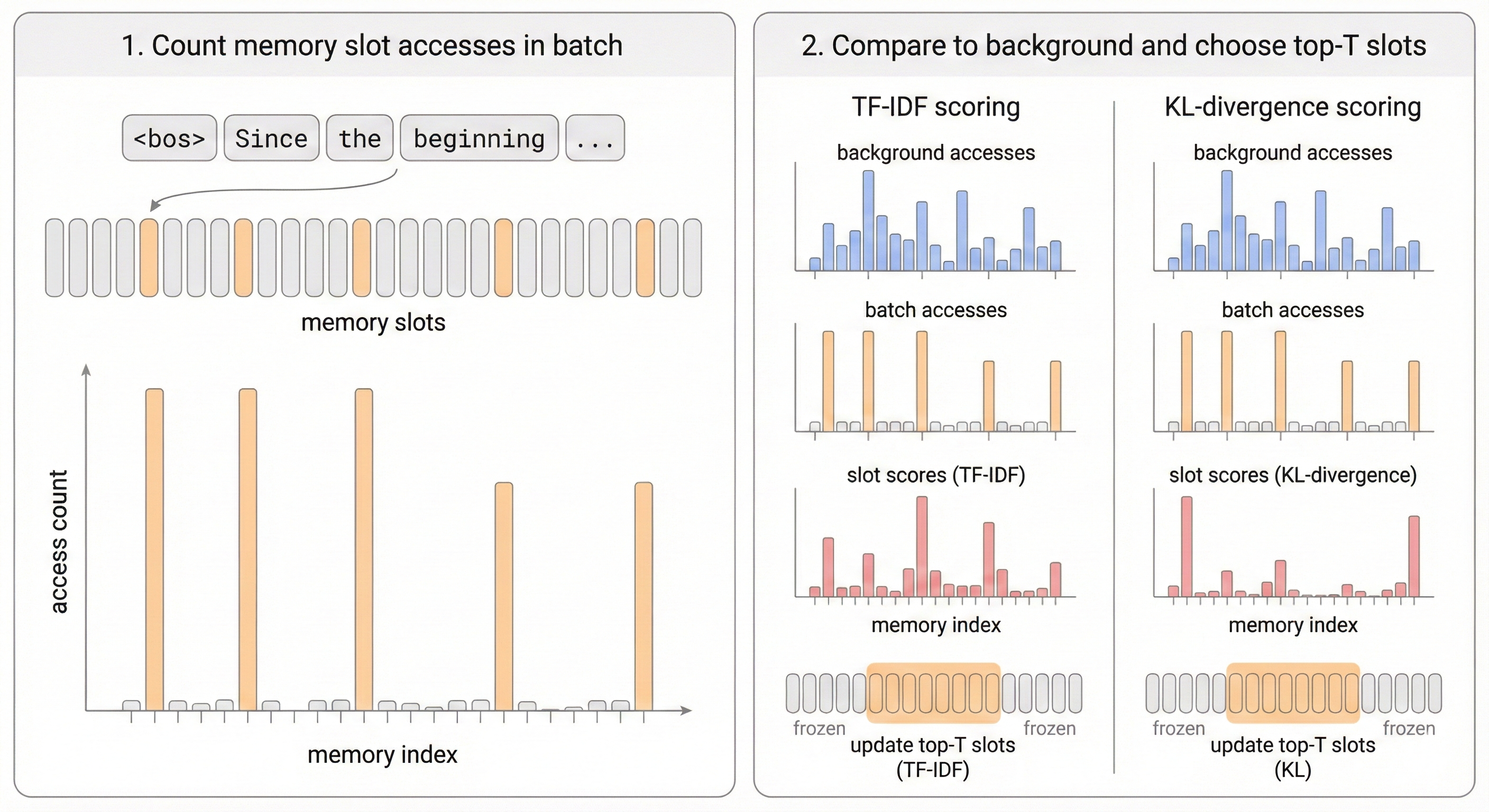}
    \caption{Sparse update mechanism: only retrieved key/value entries receive gradients.}
    \label{fig:sparseupdate}
\end{figure}

\subsection{Slot Selection for Sparse Memory Updates (TF-IDF and KL)}

A critical component of SMF is determining \textit{which} slots to update. We assume that updating slots corresponding to common, generic knowledge causes interference, while updating "surprising" or task-specific slots preserves stability. We evaluate two scoring functions to select the top-$T$ slots for updating.

We implement two slot-selection scoring rules for sparse memory updates: a TF-IDF–based baseline and a KL-divergence–based alternative, each selecting the top-$T$ accessed memory slots per batch to receive gradients. The trainer can switch between these two behaviors using a single boolean flag (kl\_div: TF-IDF when False, KL scoring when True).

\paragraph{Logging per-batch slot usage.}
For each memory layer, we attach a forward hook to the values table that records how many times each slot index is retrieved in the batch. Let $c(i)$ denote the batch count for slot $i$, and let $C=\sum_j c(j)$. Slots with $c(i)=0$ are never considered for updating in that batch.

\paragraph{Background document frequency.}
Before sparse finetuning, we compute a background statistic over $N$ batches from a background dataset (in our code: $N \le 200$ batches, batch size $1$). For each slot $i$, we compute a document-frequency-like count
\[
df(i) = \#\{\text{background batches where } c(i)>0\}.
\]
This approximates how broadly a slot is used under generic data.

\paragraph{TF-IDF slot scoring (baseline).}
For a batch, we compute term frequency $tf(i)=c(i)/C$ and inverse document frequency
\[
idf(i)=\log\frac{N+1}{df(i)+1}.
\]
We score accessed slots by
\[
s_{\text{tfidf}}(i)=tf(i)\cdot idf(i),
\]
mask out all slots with $c(i)=0$, and select the top-$T$ scoring slots.

\paragraph{KL-divergence slot scoring (our novel variant).}
We also implement an information-theoretic alternative that prioritizes slots whose usage is unexpected relative to background usage. We form the batch usage distribution
\[
p_{\text{batch}}(i)=\frac{c(i)}{C},
\]
and a smoothed background distribution derived from $df$:
\[
p_{\text{bg}}(i)=\frac{df(i)+1}{\sum_j (df(j)+1)}.
\]
We score each accessed slot by its contribution to $D_{\mathrm{KL}}(p_{\text{batch}}\|p_{\text{bg}})$:
\[
s_{\text{kl}}(i)=p_{\text{batch}}(i)\log\frac{p_{\text{batch}}(i)+\epsilon}{p_{\text{bg}}(i)+\epsilon},
\]
then select the top-$T$ accessed slots by $s_{\text{kl}}(i)$.

\paragraph{Enforcing sparse updates.}
Given the selected slot set $\mathcal{S}$ (top-$T$), we apply a gradient hook on the memory values matrix $V \in \mathbb{R}^{M\times d}$ to zero out all rows not in $\mathcal{S}$, ensuring only those slots are updated by the optimizer.

\section{Evaluation}

\subsection{Experimental Setup}
We evaluate whether the retrofitted model can learn a new task while retaining held-out knowledge.

We begin from Qwen-2.5-0.5B-Instruct and perform recovery on OpenAssistant (10k samples). After recovery, we finetune on TriviaQA (1k samples). We compare (i) a dense finetuning baseline against (ii) sparse memory finetuning where only accessed memory entries are updated. The base model is otherwise frozen during memory-focused stages, matching the goal of isolating updates.

\subsection{Metrics}

We evaluate target-task performance using F1 on TriviaQA or SimpleQA and measure forgetting on held-out tasks using GSM8K loss and Natural Questions F1. The key metric of interest is the tradeoff between target-task adaptation and retention of general knowledge: methods that improve the target task while degrading held-out performance exhibit forgetting, whereas stable methods minimize such degradation. The TriviaQA and SimpleQA settings correspond to the small-data fact-learning and document-level QA regimes studied in \cite{lin2025continuallearningsparsememory}, respectively, enabling evaluation of continual learning.

\section{Results}
Figures~4 and~5 illustrate recovery-phase behavior under two different adaptation settings, training on TriviaQA and SimpleQA respectively, while evaluating on Natural Questions and GSM8K. 

\textbf{Plasticity:} In both settings, sparse memory finetuning enables rapid adaptation to the target task, achieving strong F1 scores within a few hundred steps. Full finetuning shows comparatively limited improvement on the target task in this timeframe, suggesting that modifying the massive dense backbone requires more data or steps to converge than the lightweight memory updates.

\begin{figure}[t]
    \centering
    \begin{subfigure}{0.32\linewidth}
        \centering
        \includegraphics[width=\linewidth]{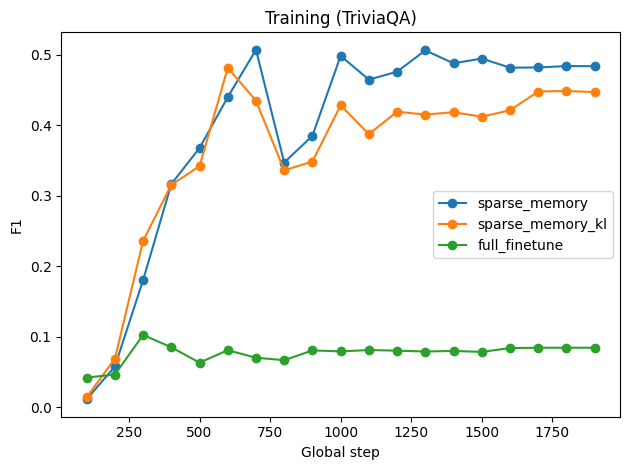}
        \caption{TriviaQA}
        \label{fig:recovery_triviaqa}
    \end{subfigure}
    \hfill
    \begin{subfigure}{0.32\linewidth}
        \centering
        \includegraphics[width=\linewidth]{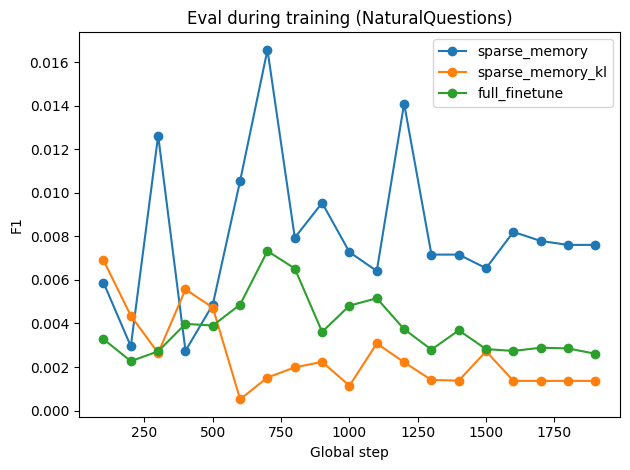}
        \caption{Natural Questions}
        \label{fig:recovery_nq}
    \end{subfigure}
    \hfill
    \begin{subfigure}{0.32\linewidth}
        \centering
        \includegraphics[width=\linewidth]{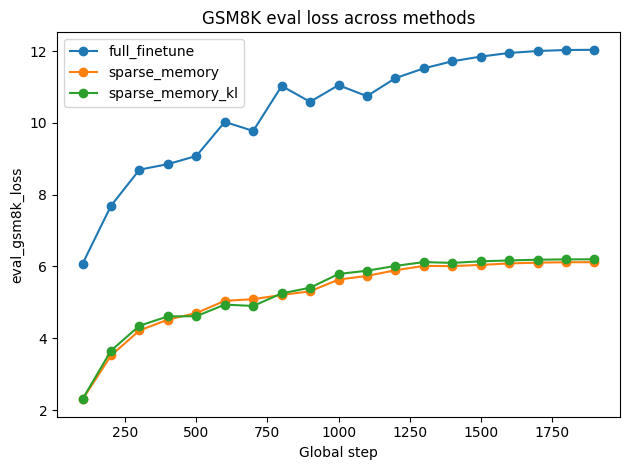}
        \caption{GSM8K}
        \label{fig:recovery_gsm8k}
    \end{subfigure}

    \caption{Finetuning performance across datasets. (a) TriviaQA training F1, (b) Natural Questions evaluation F1 during recovery, and (c) GSM8K evaluation loss for full finetuning and sparse memory variants.}
    \label{fig:recovery}
\end{figure}
\begin{figure}[t]
    \centering
    \begin{subfigure}{0.32\linewidth}
        \centering
        \includegraphics[width=\linewidth]{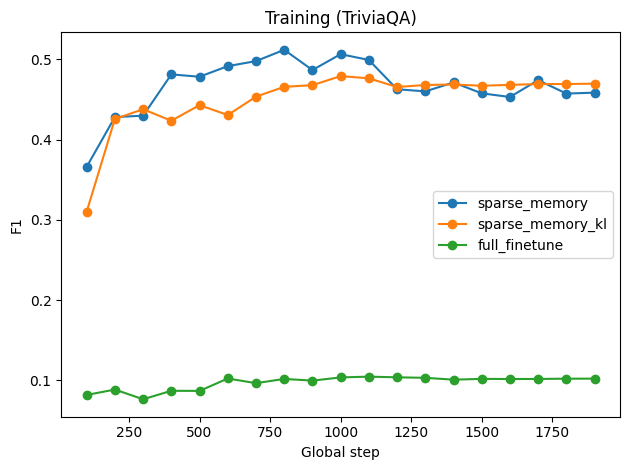}
        \caption{TriviaQA}
        \label{fig:recovery_simpleqa}
    \end{subfigure}
    \hfill
    \begin{subfigure}{0.32\linewidth}
        \centering
        \includegraphics[width=\linewidth]{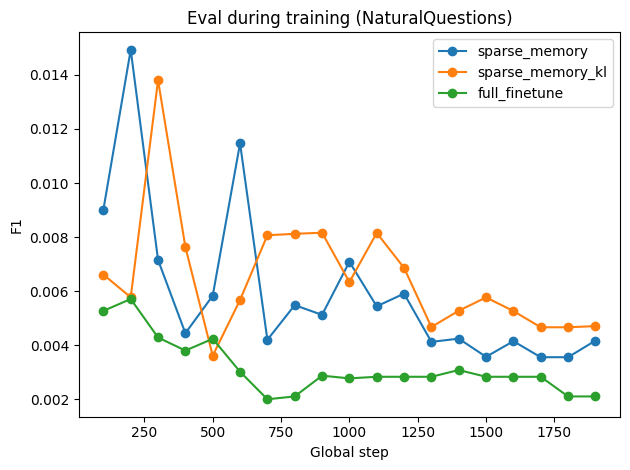}
        \caption{Natural Questions}
        \label{fig:recovery_nq2}
    \end{subfigure}
    \hfill
    \begin{subfigure}{0.32\linewidth}
        \centering
        \includegraphics[width=\linewidth]{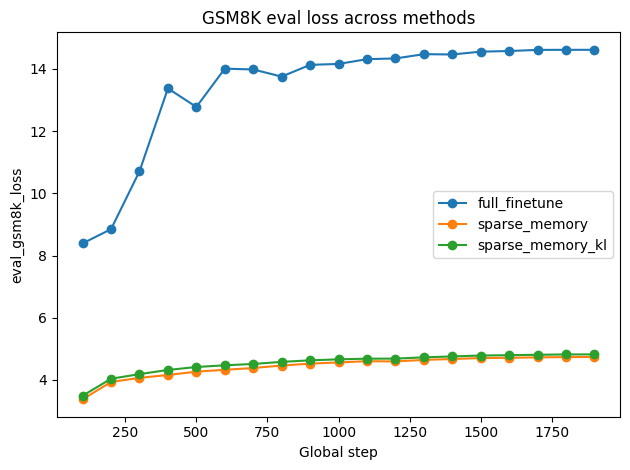}
        \caption{GSM8K}
        \label{fig:recovery_gsm8k2}
    \end{subfigure}

    \caption{Finetuning performance when adapting to SimpleQA. (a) TriviaQA training F1, (b) Natural Questions evaluation F1 during recovery, and (c) GSM8K evaluation loss for full finetuning and sparse finetuning variants.}
    \label{fig:recovery2}
\end{figure}

\textbf{Stability (Forgetting):}
The results reveal a task-dependent stability-plasticity tradeoff:
\begin{itemize}
    \item \textbf{TriviaQA:} When training on TriviaQA (Figure~\ref{fig:recovery_triviaqa}), sparse finetuning \textit{without} KL-divergence slot scoring preserves performance on NaturalQuestions more effectively. The KL-scoring variant, while stable, exhibits slightly greater forgetting in this specific regime, likely because the strong gradient signal from TriviaQA conflicts with the KL constraint to stay close to the background distribution.
    \item \textbf{SimpleQA:} In contrast, when training on the SimpleQA dataset (Figure~\ref{fig:recovery2}), KL regularization proves essential. It stabilizes learning, preventing excessive drift caused by noisy updates.
    \item \textbf{Dense Finetuning:} Across both settings, full finetuning consistently degrades performance on GSM8k (increasing loss), indicating catastrophic forgetting of reasoning capabilities.
\end{itemize}

\section{Conclusion}
We propose a continual learning framework for pretrained language models that augments transformer layers with memory layers updated via sparse finetuning. We introduce two sparse update mechanisms: TF-IDF–based slot scoring and a novel KL-divergence–based scoring method, and evaluate them against full finetuning on target tasks while monitoring performance on held-out benchmarks. Across TriviaQA and SimpleQA recovery settings, sparse finetuning enables rapid adaptation with minimal forgetting, whereas full finetuning exhibits limited learning and severe degradation on reasoning tasks such as GSM8K. KL-based scoring reveals a task-dependent stability–plasticity tradeoff, improving retention when adaptation signals are weak and underscoring the effectiveness of sparsely updated memory layers for continual learning.




    

\newpage
\printbibliography


\end{document}